\documentclass[10pt,twocolumn,letterpaper]{article} 

\usepackage{wacv}
\usepackage{amsmath} 
\usepackage{amssymb}  
\usepackage{booktabs}
\usepackage{epsfig}
\usepackage{graphicx}
\usepackage{makecell}

\usepackage[pagebackref=true,breaklinks=true,letterpaper=true,colorlinks,bookmarks=false]{hyperref}

\usepackage{color}
\usepackage{gensymb}
\wacvfinalcopy
\def\wacvPaperID{302} 

\ifwacvfinal\fi
\begin{document}

\title{Matching Disparate Image Pairs Using Shape-Aware ConvNets}

\author{Shefali Srivastava\textcolor{green}{$^{\dagger}$}$^{,1, 2}$ \vspace{-0.03 in}\\
{\tt\small shefali9625@gmail.com}
\vspace{-0.05 in}
\and
Abhimanyu Chopra\textcolor{green}{$^{\dagger}$}$^{,2}$ \vspace{-0.03 in}\\
{\tt\small abhi.livemail@gmail.com }
\vspace{-0.05 in}
\and
Arun CS Kumar$^1$ \vspace{-0.03 in} \\
{\tt\small aruncs@uga.edu}
\vspace{-0.05 in}
\and
Suchendra M. Bhandarkar$^1$ \vspace{-0.03 in} \\
{\tt\small suchi@cs.uga.edu }
\vspace{-0.05 in}
\and
Deepak Sharma$^2$ \vspace{-0.03 in} \\
{\tt\small dk.sharma1982@yahoo.com }
\vspace{-0.05 in}
\and
\vspace{-0.05 in}
\small
$^1$ Department of Computer Science, The University of Georgia, Athens, GA 30602-7404, USA\\
\small
$^2$ Department of Information Technology, Netaji Subhas Institute of Technology, New Delhi 110078, India\\
\vspace{-0.25 in}
\small{(\textcolor{green}{$\dagger$} indicates joint first authors)}
}

\maketitle

\ifwacvfinal\thispagestyle{empty}\fi
\thispagestyle{empty}
\pagestyle{empty}

\begin{abstract}
\vspace{-0.3cm}
An end-to-end trainable ConvNet architecture, that learns to harness the power of shape representation for matching disparate image pairs, is proposed. Disparate image pairs are deemed those that exhibit strong affine variations in scale, viewpoint and projection parameters accompanied by the presence of partial or complete occlusion of objects and extreme variations in ambient illumination. Under these challenging conditions, neither local nor global feature-based image matching methods, when used in isolation, have been observed to be effective. 
The proposed correspondence determination scheme for matching disparate images exploits high-level shape cues that are derived from low-level local feature descriptors, thus combining the best of both worlds. 
A graph-based representation for the disparate image pair is generated by constructing an affinity matrix that embeds the distances between feature points in two images, thus modeling the correspondence determination problem as one of graph matching. 
The eigenspectrum of the affinity matrix, i.e., the learned global shape representation, is then used to further regress the transformation or homography that defines the correspondence between the source image and target image. The proposed scheme is shown to yield state-of-the-art results for both, coarse-level shape matching as well as fine point-wise correspondence determination.

\end{abstract}

\begin{figure}
 \centering
 \vspace{-0.1cm}
  \includegraphics[width=8.5cm, height=8cm]{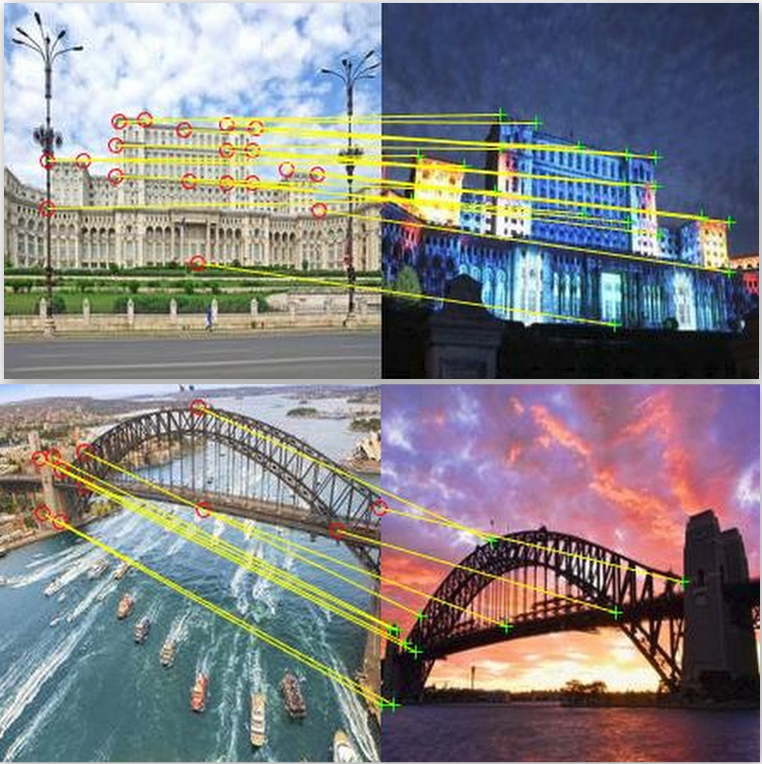}
  \vspace{-0.3cm}
  \caption{\footnotesize Examples of disparate image pairs with correspondences determined by the proposed method.}  
  \label{fig:dip}
  \vspace{-0.4cm}
\end{figure}


\vspace{-0.5cm}
\section{Introduction}
\vspace{-0.1cm}

Image matching is one of the most fundamental and well explored problems in computer vision. Significant research effort has been directed towards addressing this problem, primarily on account of its widespread and extensive applicability in several other computer vision problems such as structure-from-motion (SfM), object localization, fine-grained object categorization, content-based image retrieval (CBIR), to mention a few. Establishing correspondences across disparate images is a special case of the image matching problem where the underlying images exhibit extreme variations in scale, orientation, viewpoint, illumination and affine projection parameters accompanied by the presence of complete or partial occlusion of objects. Matching of disparate images is of particular importance, owing to its direct employability in several real-world applications such as (a) \textit{self-driving cars} and \textit{autonomous robot navigation}, where matching images from different viewpoints and under varying illumination conditions coupled with occlusion reasoning is critical for 3D scene reconstruction; (b)  \textit{virtual phototourism} where it is important to match images of buildings and monuments that are taken under varying illumination conditions (day {\it vs.} night) and from different historic periods (ancient {\it vs.} modern) for the purpose of content-based retrieval and 3D scene reconstruction; and (c) \textit{mixed/augmented reality} and \textit{virtual gaming}, where it is important to match disparate images characterized by clutter and textureless regions while ensuring reliable reconstruction of the encompassing 3D space. 

In spite of extensive research over the past three decades, the problem of disparate image matching is largely unsolved. Obtaining consistent, robust, well-grounded and definitive results while matching image pairs with disparities beyond a certain degree, is still an open challenge. What makes this problem especially challenging? For the most part, it is the presence of strong articulations, viewpoint variations, difference in illumination (day {\it vs.} night), and presence of occlusions. Real world images often tend to exhibit extreme variations in the affine transformations that describe the scales at and viewpoints from which the images are captured, variations in illumination (day {\it vs.} night), variations in structure (historic {\it vs.} new), and presence of occlusion typically caused by external objects or self-occlusion due to viewpoint changes. In the face of such large disparities, both global shape-based and local feature-based image matching methods fail to provide robust and accurate results. 

Most commonly used approaches to disparate image matching involve establishing one-to-one point correspondences or matches between local pixel-level feature descriptors, such as \textit{scale-invariant feature transform} (SIFT) keypoints~\cite{lowe2004distinctive}, extracted from the two images. The computed descriptor differences between the two images are used to generate point-wise correspondences. Although local feature-based approaches are typically fast and generally effective, they fare well only when matching images that are reasonably similar. Since local feature descriptors treat the underlying image as a spatial distribution of independent keypoints, they are unsuccessful in reasoning about the higher-level object shape cues that project onto the 2D keypoints. Since keypoint-based local descriptors are unable to capture adequately the global object-level shape, their matching performance suffers significantly when the extent of image dissimilarity is high~\cite{Bansal13}. 

An alternative school of thought is based on exploiting higher-level representations such as object shapes or structures for matching images~\cite{Belongie02,Bai07,Zhu08}. Humans often rely on global shape or semantic cues to quantify image correspondences as opposed to matching images based on local keypoint-based feature similarities. However, it should be emphasized that most global shape matching methods lack the ability to model shape articulations effectively. Thus, in our approach, we propose to compute a global shape representation via point-wise correspondences in an attempt to reason about image correspondences at a higher level. We learn to match disparate images using a global shape representation while harnessing the descriptive power of local keypoint-based features. To this end, we formulate the problem of matching shapes between disparate images as one of graph matching, and solve it using an end-to-end trainable convolutional neural network (CNN) architecture.

\begin{figure*}
 \centering
 \vspace{-0.2cm}
  \includegraphics[width=18cm, height=7cm]{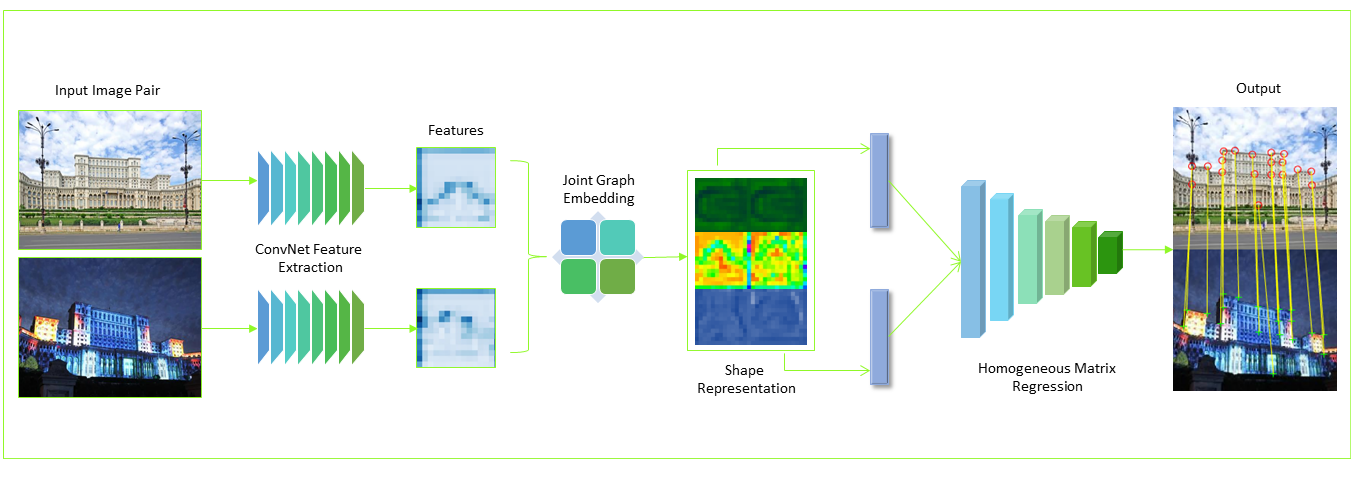}
  \vspace{-0.6cm}
  \caption{\footnotesize Pipeline of the proposed architecture: An image pair is input to a Siamese CNN to learn a feature representation which is used to compute a shape representation based on joint graph embedding, followed by a sub-network for homography matrix regression.}  
  \vspace{-0.4cm}
  \label{fig:pipeline}
\end{figure*}

The recent ubiquity of image data accompanied by rapid advances in high performance computing have unlocked the potential of machine learning using multilayer neural networks, i.e., deep learning. The availability of image data on an unprecedented scale coupled with advances in computer hardware has allowed researchers to increasingly tackle computer vision problems using data-driven approaches. This trend is reflected in the pervasive use of multilayer convolutional neural network (CNN) architectures with sophisticated structures and potent feature learning capabilities in several computer vision applications. Most importantly, the ability of CNNs to learn complex hierarchical object representations, accompanied by the backpropagation learning algorithm with a well-behaved gradient function, makes them a powerful tool in several computer vision problems. 

There have been some recent attempts at exploiting high-level object shape representations in image matching~\cite{Bansal13,Mukhopadhyay16}. However, only a few attempts have focused on deriving more meaningful low-level representations of the underlying data rather than on matching patches or regions across images~\cite{Bansal13,Mukhopadhyay16}. In this paper, we propose an novel CNN architecture, trained in an end-to-end fashion, that attempts to determine a global shape-based representation of scene structures from local features. Latent feature representations extracted from the input images using pre-trained CNN models, are used to compute a joint graph that is represented by a suitably defined affinity matrix. The affinity matrix jointly embeds the feature similarity between the input images. The eigenvectors and eigenvalues resulting from the spectral decomposition of the affinity matrix are deemed to be the global shape representation. The eigenvectors are used to learn to regress the homography that minimizes a point-wise correspondence loss function and to train the network to learn to reason about image matching via the computed shape representation. While the eigenvalues and eigenvectors obtained via eigenspectral decomposition of the affinity matrix are shown to reliably represent the underlying global object shape~\cite{Mukhopadhyay16}, reasoning about object shapes purely as linear combinations of the eigenvectors (using the eigenvalues as coefficients) is insufficient to capture the complexities of real-world shapes, especially in the face of large image disparities. The goal of this work is to relax the linearity constraints and introduce nonlinearity within the shape eigenspace using CNNs. This would allow one to reason about shapes and their variations as nonlinear combinations of the shape eigenbasis vectors (i.e., eigenvectors of the affinity matrix).

The primary contributions of this paper are: (1) formulation of an end-to-end shape-aware CNN architecture that learns to determine the point-wise correspondences needed to match disparate image pairs, and (2) formulation of a deep learning-based solution to the graph matching problem via eigenspectral decomposition. We have evaluated the proposed scheme on three challenging datasets~\cite{oxfordvgg,hpatches_2017_cvpr,cskumar2018}, for both shape matching as well as point-wise correspondence estimation tasks and achieved competitive measures of performance. We have successfully demonstrated the ability of the proposed CNN architecture to perform matching on images from previously unseen datasets~\cite{oxfordvgg,hpatches_2017_cvpr}, while being trained only on the dataset of~\cite{cskumar2018}. This justifies the feasibility of the proposed scheme as an off-the-shelf image matching technique.

\vspace{-0.1cm}
\section{Related Work}
\vspace{-0.15cm}
Image matching techniques in the research literature can be broadly categorized as global shape-based techniques~\cite{Belongie02,Bai07,Zhu08} or local point-based~\cite{Chen08,Ma11} techniques. Global shape-based matching techniques rely on extracting the overall shape of an object (or structure) within the image. High-level shape cues are extracted from the underlying images to compute a degree of similarity. Global shape-based matching techniques are further subclassified as region-based~\cite{Kim00,Sebastian04} or contour-based~\cite{Bai07,Zhu08,Opelt06}. Contour-based methods exploit peripheral information to augment the underlying shape-based features. Shape skeleton-based contour matching~\cite{Bai07} and dynamic programming~\cite{Ravishankar08} are used to compute a similarity measure such as shape context~\cite{Belongie02}, chamfer distance~\cite{Opelt06} or set matching-based contour similarity~\cite{Zhu08}. Other contour-based matching methods include segment-based matching techniques such as hierarchical Procrustes matching~\cite{Mcneill06}, shape tree-based matching~\cite{Falzenszwalb07} and triangle area-based matching~\cite{Alajlan07}. However, the aforementioned contour-based matching techniques fall short when dealing with significant articulations in the object shapes. Global region-based matching approaches characterize the underlying shapes using global descriptors such as Zernike moments~\cite{Kim00} which are invariant to affine transformations. Skeleton-based shape descriptors~\cite{Sebastian01,Sebastian04} are better at capturing shape articulations, but their performance diminishes with increasing shape complexity. 

Local point-based matching methods rely primarily on local appearance-based descriptors to identify correspondences. They attempt to tackle shape deformation by deconstructing the constellation of local appearance descriptors~\cite{Chen08,Ma11}. The local feature points are identified by extracting salient image regions and combined to create meaningful shape representations. While local feature descriptors provide a reliable measure of local similarity, they are limited in their ability to estimate accurate shape correspondence. Also, since most local point-based matching techniques rely on prior knowledge of the underlying shape, their applicability to real-world images is limited~\cite{Chen08,Ma11}.  

Both, local point-based and global shape-based image matching approaches have their advantages and shortcomings. While global shape-based descriptors are well behaved and less sensitive to outliers, they, in isolation, are insufficient to compute point-wise correspondences since they do not explicitly encode keypoint information as appearance-based descriptors do~\cite{Bansal13,Mukhopadhyay16}. Although global shape descriptors are shown to perform well when the image pairs exhibit significant shape deformations due to changes in viewpoint, their performance suffers in the presence of strong shape articulations~\cite{Mukhopadhyay16}. Region-based global descriptors are also vulnerable to instances of partial occlusions. While global shape representations have the advantage of being able to employ global shape cues for matching, their computational complexity and inability to compute accurate point-wise correspondences makes them unsuitable for most practical applications that demand reliable point-wise correspondences. In contrast, local point-based descriptors, in theory, are capable of yielding more reliable keypoint correspondences, but are often plagued by noisy matches and can also prove to be computationally expensive.

In recent times, several attempts have been made to leverage the representational power of deep learning methods to improve image matching~\cite{Zagoruyko15,detone2017toward,dosovitskiy2015flownet}. Zagoruyko {\it et al.}~\cite{Zagoruyko15} have explored a variety of CNN architectures to learn a similarity function for unsupervised matching of image patches. More recently, deep learning-based architectures have been proposed to predict and identify SIFT feature-like keypoints for incorporation into traditional SfM pipelines~\cite{detone2017toward,simo2015discriminative}. Simo {\it et al.}~\cite{simo2015discriminative} learn discriminant patch representations using a Siamese CNN architecture to identify and represent keypoints. DeTone {\it et al.}~\cite{detone2017toward,detone2017superpoint} have proposed a self-supervised CNN architecture that learns keypoints from single images by warping images to known transformations, thereby rendering image pairs for supervised learning. Yi {\it et al.}~\cite{yi2016lift} use SfM reconstructions for supervised learning and prediction of keypoints by a Siamese CNN. 

As an alternative to keypoint-based schemes, optical flow-based methods have been proposed to learn to regress correspondences as a continuous function in a data-driven manner~\cite{dosovitskiy2015flownet, mayer2016large, zhou2017unsupervised}. Dosovitskiy {\it et al.}~\cite{dosovitskiy2015flownet} first demonstrated that optical flow could be formulated as a supervised learning problem, and their work was further improved upon by Mayer {\it et al.}~\cite{mayer2016large} to predict high-resolution optical flow maps. Zbontar {\it et al.}~\cite{zbontar2016stereo} have used stereo pairs for supervised training of a CNN for learning patch-wise similarity/dissimilarity to compute disparity maps, whereas~\cite{zhou2017unsupervised, kumar2018monocular} use image consistency loss to unsupervisedly learn optical flow from videos. Some techniques have attempted to search for image patches and leverage the representational features of patches to improve optical flow computation~\cite{seki2016patch, mayer2016large}.

Unlike keypoint-based approaches, optical-flow based approaches lack the ability to model shape articulations. They also require vast amounts of training data and assume temporal supervision along with prior knowledge of frame rates. In contrast, keypoints can be discovered with relatively minimal supervision~\cite{Zagoruyko15,detone2017toward,yi2016lift} even from single images~\cite{detone2017superpoint}. Keypoint-based approaches are capable of reliably generating fine correspondences, making them better suited for SfM pipelines. However matching of keypoints and subsequent regularization of matches are expensive and ensuring end-to-end differentiability for SfM pipelines is challenging. While both approaches learn in a data-driven manner while maintaining end-to-end differentiability, they learn to simply regress from within the image/tensor space, thereby failing to exploit the underlying shape representation.

The proposed scheme aims to overcome the above shortcomings, by integrating both global and local methods within an end-to-end trainable deep learning framework. Inspired by~\cite{Bansal13}, we leverage the representational power of deep learned feature descriptors to construct a graph representation characterized by an affinity matrix. Spectral decomposition via joint diagonalization of the affinity matrix, allows a high-level shape representation based on the eigenvectors and eigenvalues of the affinity matrix. The computed shape representation is used to regress the final homography matrix via an independent CNN sub-network. The proposed scheme is shown to be very effective in matching global image descriptors as well as estimating point-wise correspondences. 

\begin{table*}[h]
\vspace{-0.3cm}
\scriptsize
	\renewcommand{\arraystretch}{0.95}
	\centering
    \setlength{\tabcolsep}{10pt}
	\caption{\footnotesize Architectural details of the proposed CNN}
    \vspace{0.05 in}
	\label{tab:network_arch}
	\begin{tabular}{llrrcr}
    	\toprule
		Phase 	&	Layer & Input Size & Filter Size & No. of Filters & Output Size\\ \midrule 
		Stage 1: Siamese Network & Input Layer & $224 \times 224 \times 3$ & -Multiple- & -Multiple- & $28 \times 28 \times 512$ \\
        Stage 2: Joint Graph Embedding & JGGE & ($\mathcal{F}_1$) $28 \times 28 \times 512$& - & - & \\
        &  & ($\mathcal{F}_2$) $28 \times 28 \times 512$ & - & - & $28 \times 28 \times 2m$\\
        Stage 3: Regress Homography Matrix& Conv11 & $28 \times 28 \times m$ & $3 \times 3$ & $128$ & $28 \times 28 \times 128$ \\
        & Conv12 & $28 \times 28 \times m$ & $3 \times 3$ & $128$ & $28 \times 28 \times 128$ \\
        & ConvMerge & $28 \times 28 \times 128$ & - & - & - \\
        & & $28 \times 28 \times 128$ & - & - & $28 \times 28 \times 256$ \\
        & Conv2 & $28 \times 28 \times 256$ & $3 \times 3$ & $96$ & $28 \times 28 \times 96$ \\
        & MaxPool2 & $28 \times 28 \times 96$ & $2 \times 2$ & $-$ & $14 \times 14\times96$ \\
        & Conv3 & $14 \times 14 \times 96$ & $3 \times 3$ & $48$ & $14 \times 14 \times 48$ \\
        & MaxPool3 & $14 \times 14 \times 48$ & $2 \times 2$ & $-$ & $7 \times 7 \times 48$ \\
        & Dense1 & $7 \times 7 \times 48$ & - & - & $256$\\ 
        & Dense2 & $256$ & - & - & $128$\\
        & Dense3 & $128$ & - & - & $9$\\
        \bottomrule
	\end{tabular}
    \vspace{-0.5cm}
\end{table*}

\vspace{-0.1cm}
\section{Motivation}
\vspace{-0.15cm}
Real-world images, such as those acquired from the Internet, often tend to be disparate, displaying a plethora of variations as shown in Figure~\ref{fig:dip}. Most local keypoint-based reconstruction methods fail at reliable matching of Internet-acquired images on account of their errant nature. To ensure reliable image matching, local keypoint-based reconstruction techniques are compelled to use large training image datasets to reconstruct scenes that would otherwise require significantly fewer training images. 

To address matching inaccuracies induced by the image disparities, we resort to higher-level cues such as the object shape. Shape cues are more resilient to illumination changes and affine transformations, but fail in the presence of occlusion and strong shape articulation. While shape articulation is minimal in the case of rigid scene structures (i.e., the objects of interest in this paper), instances of full or partial occlusion are quite common. In contrast, local keypoint-based methods are more robust to partial occlusion. The natural question then arises \textemdash can we combine both?

While there have been several attempts to combine local and global cues for image matching, we draw our inspiration from~\cite{Bansal13,Mukhopadhyay16} which show that a higher-level shape representation computed via eigenspectral analysis of a low-level feature representation, can yield matching features that are persistent across illumination changes and other affine variations~\cite{Bansal13}. In particular, features that encode the extrema of the eigenfunctions of the joint image graph are shown to be stable, persistent and robust across wide range of illumination variations~\cite{Bansal13,Mukhopadhyay16}. While the proposed hybrid approach combines both global and local cues efficiently, a key question remains, i.e., how does one extend the approach to enable end-to-end learning in a data-driven manner? In response to this question, we propose a novel CNN architecture that learns to identify features and reason about shape in a suitably defined eigenspace to further regress a homography, while being trained end-to-end.

\vspace{-0.1cm}
\section{Proposed Model}
\vspace{-0.15cm}

We construct an end-to-end trainable CNN architecture comprising of (1) a {\it Siamese} network, encompassing a pair of standard VGGNet CNNs~\cite{simonyan2014very} truncated up and until the fourth convolutional layer ({\it conv\_}$4$) with pretrained weights, that output a pair of tensors for each image, (2) a joint graph formulation computed as shown in equation~(\ref{eq:EMBD}), which is then reshaped into 2D grids of corresponding sizes and further appended with (3) a pair of (Siamese) convolutional layers, which are then concatenated, followed by a few dense layers to ultimately regress the homography matrix $H$. The details of the network architecture are presented in Table~\ref{tab:network_arch}.  

\vspace{-0.0cm}
\subsection{Siamese Network: Feature Learning}
\vspace{-0.1cm}
We use a pair of VGGNet CNNs\cite{simonyan2014very} pretrained on the Imagenet dataset~\cite{russakovsky2015imagenet} to construct a Siamese network with shared weights. We truncate the individual VGGNets at the fourth convolutional layer ({\it conv\_4}). The {\it conv\_4} features from the two images $I_1$ and $I_2$ are represented as $\mathcal{F}_1$ \& $\mathcal{F}_2 \in \mathbb{R}^{N\times 512} $ respectively, in all further discussions. We use the features $\mathcal{F}_1$ \& $\mathcal{F}_2$ to compute the joint graph embedding as explained in Section~\ref{joint_graph}. Also, we remove the max pooling layers in the default VGGNet, and instead increase the stride parameters of the convolutional layer to downsize the tensor. Although the max pooling layer is useful for most discriminative tasks, it forces the network to learn spatial location invariance, which we would prefer to avoid in the proposed scheme. 

\begin{figure*}
\vspace{-0.2cm}
 \centering
  \includegraphics[width=\textwidth]{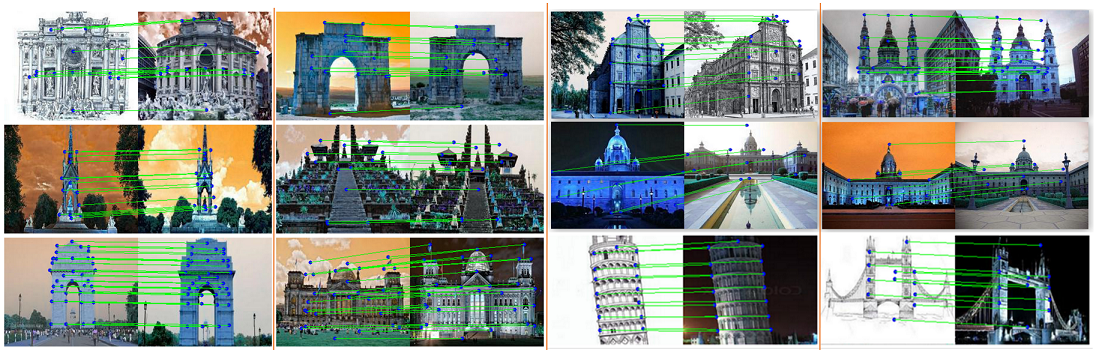}
  \caption{Qualitative results on the {\it DispScenes} dataset: Images from the test set of the \textit{DispScenes} dataset~\cite{cskumar2018} with estimated point-wise correspondences plotted as lines across the image pairs.}  
  \vspace{-0.4cm}
  \label{fig:qual1}
\end{figure*}

\vspace{-0.0cm}
\subsection{Shape Network: Joint Graph Embedding}\label{joint_graph}
\vspace{-0.1cm}
A weighted image graph $G(V,E,W)$ is computed from the convolutional layer features represented as $\mathcal{F}_1$ \& $\mathcal{F}_2 \in \mathbb{R}^{N\times 512} $ derived from images $I_1$ and $I_2$ respectively. In $G(V,E,W)$, the set $V$ denotes the vertices of the graph. The pairwise feature distances, represented by the edge set $E$, model the pair-wise relationships between vertices in $V$. The distance matrix $W$, an adjacency matrix that represents the graph, is formulated as shown in the equation~(\ref{eq:EMBD}). For every edge $(v_i,v_j) \in E$ the weight $w_{ij} \geq 0$ encodes the affinity between pixel-level features of vertices $v_i$ and $v_j$. If the number of points in the image are $n$, the affinity matrix $W$, as computed above, is of size $n \times n$ and represented as $W = [w_{ij}]_{i,j=1,2..,n}$

The above formulation is further extended to construct a joint representation that embeds the global dissimilarity encoding the differences between the features of the two images. Consider the image graphs $G_1(V_1,E_1,W_1)$ and $G_2(V_2,E_2,W_2)$ of the two images $I_1$ and $I_2$ respectively. A joint graph $G(V,E,W)$ is defined by the vertex set $V = V_1 \cup V_2$ and edge set $E = E_1 \cup E_2 \cup (V_1 \times V_2)$ where the Cartesian product $V_1 \times V_2$ denotes the set of edges connecting every pair of vertices  $(v_i, v_j): v_i \in V_1, v_j \in V_2$ in the graph. The affinity matrix $W$ is given by:
\vspace{-0.15cm}
\begin{equation}
\vspace{-0.1cm}
   \mathbf{W}=
  \left[ {\begin{array}{cc}
   W_1 & C \\
   C^T & W_2 \\
  \end{array} } \right]_{(n_1+n_2) \times (n_1+n_2)}   \label{eq:EMBD}  
\end{equation}
\noindent where the affinity submatrices $W_1$, $W_2$ and $C$ are defined as follows:
\vspace{-0.15cm}
\begin{eqnarray}
(W_i)_{x,y} & = & \exp( -  (\|{f_i(x) - f_i(y)}\|)^2 ) \\
C_{x,y} & = & \exp( -  (\|{f_1(x) - f_2(y)} \|)^2) 
\vspace{-0.1cm}
\end{eqnarray}
where $f_i(x)$ and $f_i(y)$ are pixel-level features at locations $x$ and $y$ respectively in image $I_i$ and $\| \cdot \|$ is the cosine distance between them.

The first $m$ non-trivial eigenvectors, that correspond to the topmost $m$ non-trivial eigenvalues, are extracted from the eigenspectrum of the joint matrix $W$. Bansal {\it et al.}~\cite{Bansal13} showed that the eigenspectrum of the joint matrix, as computed above, results in reliable shape correspondences. The Joint Graph Embedding (JGE) process is shown to reduce the divergence between features derived from the corresponding regions of the two images. Consequently, regions from both images that exhibit strong correspondence in JGE space are noted to be in visual agreement with the correspondence results in image space~\cite{Bansal13}. 

\vspace{-0.0cm}
\subsection{Regression of the Homography Matrix}
\vspace{-0.1cm}
The top-$m$ eigenvectors computed in Section~\ref{joint_graph}, are fed to a cascade of convolutional and dense layers followed by max pooling layers to induce non-linearity (Figure \ref{fig:pipeline}). Only the non-trivial or top-$m$ eigenvectors are considered since they encode rich global shape information whereas eigenvectors that are trivial or outside the top-$m$ preserve mostly local cues which we deem uninformative and noisy~\cite{Mukhopadhyay16} and hence detrimental to the overall learning. Finally, the network outputs $9$ values, which are then reshaped into a $3\times 3$ grid that constitutes the homography matrix denoted as $H$. The use of the homography matrix has an added advantage as they project the points to target in a regularized manner.

The network architecture detailing the regression layers is presented in Table~\ref{tab:network_arch}. The network is trained to minimize the loss $\mathcal{L}$ given in equation~(\ref{eq:loss}):
\vspace{-0.15cm}
\begin{equation}\label{eq:loss}
\vspace{-0.1cm}
\mathcal{L} =\frac{1}{n} {\sum_{i=0}^{n} ||(P_{1}^i \times H - P_{2}^{i})||^2}
\vspace{-0.1cm}
\end{equation}
Loss $\mathcal{L}$ is the average of $L_2$ distances between the estimated and ground truth keypoints, where $n$ is the number of ground truth keypoints, $P_{1}$ and $P_{2}$ represent the annotated ground truth keypoints of images $I_1$ and $I_2$ respectively and homography matrix $H \in \mathbb{R}^{3 \times 3}$ is the output of the network described above. By minimizing the point-wise loss, the network learns to regress the homography matrix $H$ from the shape space.

\begin{figure*}
\vspace{-0.25cm}
 \centering
  \includegraphics[width=\textwidth,height=3cm]{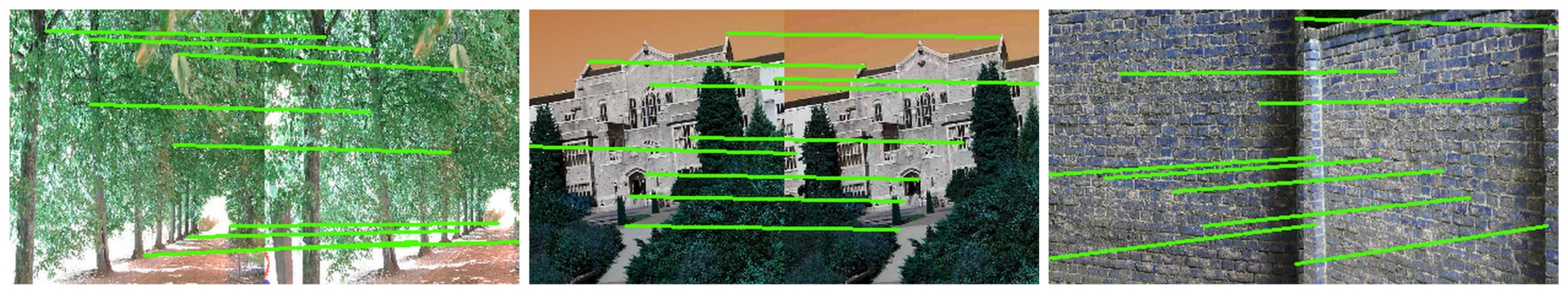}
  \vspace{-0.6cm}
  \caption{\footnotesize Qualitative results on the {\it Affine Invariant Features Oxford} dataset: Correspondences estimated by the proposed method on images from the Oxford dataset~\cite{oxfordvgg}.}  
  \label{fig:oxford}
  \vspace{-0.2cm}
\end{figure*}

\begin{figure*}
 \centering
  \includegraphics[width=\textwidth, height=3cm]{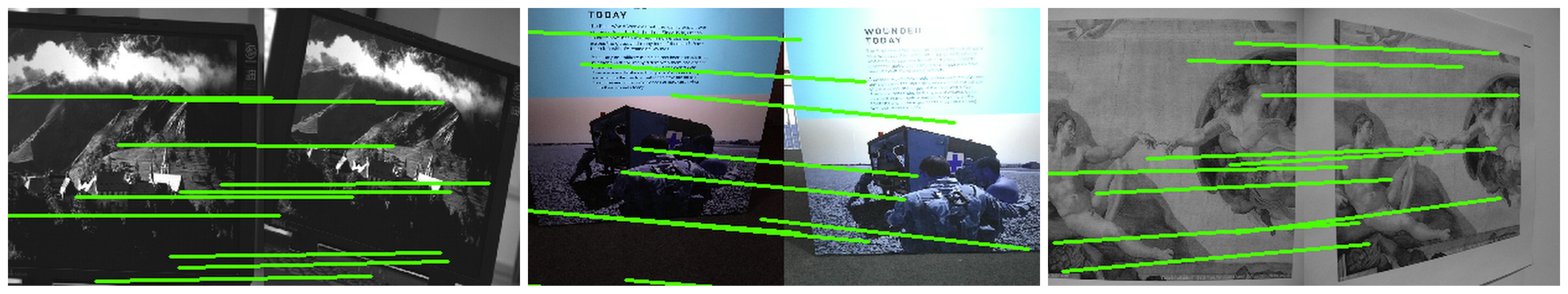}
  \vspace{-0.5cm}
  \caption{\footnotesize Qualitative results on the {\it HPatches} dataset: Correspondences estimated by the proposed method on images from the \textit{HPatches} dataset~\cite{hpatches_2017_cvpr}.}  
  \label{fig:hpatches}
  \vspace{-0.15cm}
\end{figure*}

\begin{table*}[!htbp]
\caption{\footnotesize Comparison of the point-wise correspondences estimated using the proposed scheme with those obtained via standard keypoint-based~\cite{Lowe04} techniques, deep learning features ({\it VGG-16} trained on {\it ImageNet}), and other shape matching methods such as~\cite{Bansal13,Mukhopadhyay16} and {\it end-to-end} trained deep learning methods such as~\cite{cskumar2018,detone2017superpoint}, on the \textit{DispScenes} dataset~\cite{cskumar2018}.}
 \label{tab:Compare2} 
\centering
\footnotesize
    \begin{tabular}{ c | c | c | c | c | c | c}
    \toprule

    \ & Proposed & SIFT~\cite{Lowe04} & Mukhopadhyay {\it et al.}~\cite{Mukhopadhyay16} & ConvNet ({\it VGGNet features)} & Kumar {\it et al.}~\cite{cskumar2018} & SuperPoints~\cite{detone2017superpoint}\\ \hline      
    MAE & {\bf 19.59} & 131.36 & 102.81 & 92.38 & 39.44 & 69.72 \\

\bottomrule
    \end{tabular} 
    \vspace{-0.4cm}
\end{table*}

\begin{table*}[tbp]
\vspace{-0.25cm}
\footnotesize
\caption{\footnotesize Comparison of the proposed {\it end-to-end} image matching scheme on a point-wise correspondence estimation task with approaches such as~\cite{Mukhopadhyay16} and~\cite{Hauagge12} on the \textit{DispScenes } dataset~\cite{cskumar2018}. The rows of the table represent the varying granularity $\delta$ {\it (in pixels)} within which the matched keypoints lie in proximity to the ground truth (target) keypoint. For example, given a $10 \times 10$ window centered around the target keypoint, the distance between the target and estimated ($\delta$) keypoints  must be $\leq$ 5 pixels for a valid match.}
 \label{tab:Compare3} 
\centering
    \begin{tabular}{ c | c | c | c | c | c }
    \toprule
    \ & Proposed &   DeTone et al (2017)~\cite{detone2017superpoint} & {\it Conv. features} & Mukhopadhyay et al (2016)~\cite{Mukhopadhyay16} & Bansal et al (2013)~\cite{Bansal13} \\ \hline    
    {\it $\delta \le 5$ } & 2.94 & 5.45 & 4.84 &  2.10 & 1.86\\ 
    {\it $\delta \le 10$} & 11.70 & 10.21 & 8.09 & 4.81  & 3.42\\ 
    {\it $\delta \le 15$} & 22.58  & 19.85 & 10.92 & 7.45  & 6.71\\
    {\it $\delta \le 20$} & 35.18 & 32.18 & 14.21 & 10.68 & 9.54\\ 
   
    \bottomrule
    \end{tabular} 
    \vspace{-0.2cm}
\end{table*}

\begin{table}[tbp]
\footnotesize
\caption{\footnotesize  Comparison of the proposed {\it end-to-end} image matching technique on the \textit{Oxford VGG Affine}~\cite{oxfordvgg}, \textit{DispScenes} ~\cite{cskumar2018} \& \textit{HPatches}~\cite{hpatches_2017_cvpr} datasets.}
 \label{tab:Compare4} 

    \begin{tabular}{ c | c | c | c  }
    \toprule
    \ & \textit{Oxford VGG}~\cite{oxfordvgg} & \textit{DispScenes}~\cite{cskumar2018} & \textit{HPatches}~\cite{hpatches_2017_cvpr} \\ \hline    
    {\it $\delta \le 5$} & 25.56 & 2.94  &   5.80 \\ 
    {\it $\delta \le 10$} & 46.03 & 11.70  &  23.04 \\ 
    {\it $\delta \le 15$} & 53.61 & 22.58   & 36.28 \\
    {\it $\delta \le 20$} & 62.82 & 35.18  & 46.57 \\ 
    {MAE} & 10.26 & 19.59 & 42.85 \\
   
    \bottomrule
    \end{tabular} 
    \vspace{-0.6cm}
\end{table}

\vspace{-0.1cm}
\section{Dataset}
\vspace{-0.1cm}
We use publicly available datasets~\cite{oxfordvgg,hpatches_2017_cvpr,cskumar2018} to demonstrate the performance of the proposed scheme. We train on the proposed scheme on the {\it DispScenes} dataset~\cite{cskumar2018}, the largest of all, comprising of 1029 image pairs of architectural structures. The {\it DispScenes} dataset was created to address the specific problem of disparate image matching. The image pairs in all the datasets exhibit high levels of variation in illumination and viewpoint and also contain instances of occlusion. The {\it DispScenes} dataset provides manual ground truth keypoint correspondences for all images, whereas the datasets in~\cite{oxfordvgg} and \cite{hpatches_2017_cvpr} provide the homography matrices directly. While training on the {\it DispScenes} dataset, we split the images into training, test and validation sets, where we use 70\% of the images for training, 10\% for validation and 20\% for testing. Also, while training we augment the training data with left-right flipped images generated from the corresponding raw images. Both the raw and flipped training images are subject to rotations by $90 \degree$, $180 \degree$ and $270 \degree$. Note that the annotated keypoints are subject to the same transformations as the original images.

\vspace{-0.1cm}
\section{Implementation Details}
\vspace{-0.0cm}
The first stage of the proposed network architecture consists of a Siamese network comprising of two VGGNets~\cite{simonyan2014very} with shared pretrained weights~\cite{russakovsky2015imagenet} truncated at the fourth convolutional layer. The second stage computes the eigenspectral representation detailed in Section~\ref{joint_graph}. We select the top-$m$ dominant eigenvectors (where $m=30$) of the joint matrix, extract parts of these eigenvectors corresponding to the individual images in a manner similar to~\cite{Mukhopadhyay16}, and reshape them into a pair of 28$\times$28 matrices to be passed as input to the third stage. The third stage is comprised of a series of convolutional, max pooling and dense layers that outputs an array of size 9, which is reshaped into a 3$\times$3 homography matrix. The network architecture for each stage is detailed in Table~\ref{tab:network_arch}. We chose a value of $m=30$ since it was experimentally observed to minimize the loss $\mathcal{L}$ (equation~(\ref{eq:loss})) for the validation set. The network was trained in a supervised manner to minimize the loss $\mathcal{L}$   (equation~(\ref{eq:loss})). For training, we resized input images to a resolution of 224 $\times$ 224 pixels; a standard in most object detection pipelines. The learning rate was set to 0.001 with a decay of 0.99 every 1000 steps. The network was trained on an Nvidia GeForce 1080ti GPU with 8GB RAM for 40 epochs.

\begin{figure*}

 \centering
  \includegraphics[width=\textwidth]{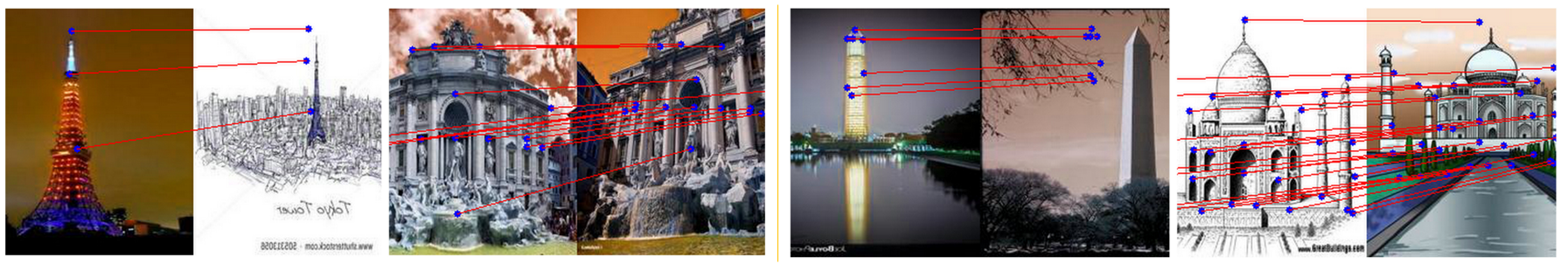}
  \vspace{-0.6cm}
  \caption{\footnotesize Qualitative results: Images from the \textit{DispScenes} dataset~\cite{oxfordvgg} with estimated point-wise correspondences displaying where the proposed scheme method failed because of the homography constraints. The point-wise correspondences estimated between image pairs show the proposed scheme to be capable of reasoning about the shape since the correspondences from the source image project well onto the target image while preserving both, the scale and shape. However, over regularization by the homography constraints results in very inaccurate point-wise correspondences.}  
  \vspace{-0.3cm}
  \label{fig:bad}
\end{figure*}

\section{Performance Evaluation}
\vspace{-0.1cm}
We evaluated the performance of the proposed scheme on the datasets in~\cite{oxfordvgg,hpatches_2017_cvpr,cskumar2018}, for both coarse shape-based and fine-grained point-wise correspondence estimation tasks. The coarse shape-based correspondence is evaluated using the mean average error (MAE) between estimated and ground truth 2D points. The MAE also quantifies the overall shape estimation performance. The MAE is computed using the homography matrix estimated by the network. The use of the homography matrix naturally smooths the correspondences and eliminates the need for {\it RANSAC}-based regularization. The results of the proposed coarse shape-based correspondence using the MAE metric is shown in Table~\ref{tab:Compare2}, alongside recent shape matching methods~\cite{Mukhopadhyay16,cskumar2018}, traditional keypoint-based methods (SIFT~\cite{lowe2004distinctive}) and recent end-to-end trained deep learning methods~\cite{detone2017superpoint}. To ensure a fair comparison, we also report matching performance results using raw {\it conv\_4} features as a baseline. 

The point-wise correspondence estimation metric used to evaluate the fine-grained matching performance of the proposed scheme is based on the extent to which the matched keypoints lie in spatial proximity of the ground truth keypoints. We compute the percentage of keypoints that lie within a $v$-pixel neighborhood given the total number of keypoints. We use varying values of $v$=$\{$5, 10, 15, 20$\}$ (in pixels), to demonstrate the matching accuracy at varying levels of granularity. For example, if $v=5$, then a match is said to be valid if the estimated keypoint lies within $5$ pixels from target (ground truth) point and within the $10 \times 10$ pixels window centered at the target point. Table~\ref{tab:Compare3} shows the performance evaluation results of the proposed scheme for fine-grained matching. In  Table~\ref{tab:Compare3}, we evaluate the proposed scheme against recent shape matching methods~\cite{Bansal13,Mukhopadhyay16,detone2017superpoint} and our baseline ({\it conv\_4} features).

We also evaluated the proposed model trained on the {\it DispScenes} dataset~\cite{cskumar2018} on the datasets in~\cite{oxfordvgg,hpatches_2017_cvpr} and reported the results in Table~\ref{tab:Compare4}. Despite the challenge of dataset bias in domain adaptation, Figures~\ref{fig:oxford} and~\ref{fig:hpatches} demonstrate that the proposed scheme can perform reliable matching on images even from unseen datasets. These results show that the proposed scheme does not overfit  the dataset, rather it learns to comprehend the shape representations accurately. Figure~\ref{fig:bad} shows cases where the proposed scheme fails due to the homography constraints, where despite being able to reason well about shape and scale, the over-regularization by the homography matrix results in large offsets in the estimated correspondences, causing the proposed scheme to produce inaccurate results.

\vspace{-0.1cm}
\section{Discussion}
\vspace{-0.1cm}
From Tables~\ref{tab:Compare2} and~\ref{tab:Compare3}, the superior performance of the proposed scheme for both, coarse shape-based and fine point-wise correspondence estimation tasks is evident. While the proposed scheme clearly outperforms traditional feature matching-based methods, it also outperforms recent shape-based methods~\cite{Mukhopadhyay16} and deep learning-based methods~\cite{cskumar2018,detone2017superpoint}. Additionally, the matching results using the baseline {\it conv\_4} layer features show that the performance improvement is not due to the choice of features, but due to the ability of the proposed scheme to learn the underlying shape representations. 

Some of the performance improvements based on the MAE metric (Table~\ref{tab:Compare3}) can be partially attributed to the use of the homography matrix, which serves as a regularizer. The other methods in Table~\ref{tab:Compare2} suffer due to the absence of a regularizer, since the MAE is impacted by the presence of outliers. 
While regularization via the homography matrix aids in minimizing the MAE in Table~\ref{tab:Compare2}, it adversely affects the performance of the proposed scheme for fine-grained correspondence estimation. The homography matrix enforces smoothness constraints, which result in approximate coarse shape matching but highly inaccurate point-wise correspondences (Figure~\ref{fig:bad}). In Figure~\ref{fig:bad}, the homography matrix estimates the affine parameters well, maintaining overall shape similarity, but the points are shifted significantly due to the enforced smoothness constraints. This accounts for the inferior performance of the proposed scheme compared to the baseline for smaller values of $v$, but superior performance for larger neighborhoods (Table~\ref{tab:Compare3}).

\vspace{-0.1cm}
\section{Conclusions}
\vspace{-0.1cm}
We proposed an {\it end-to-end} shape-aware CNN architecture for matching disparate images using joint graph embedding, where we inferred high-level (global) shape features via eigenspectral analysis of low-level (local) image features. We posed the problem of image matching as a graph matching problem and solved it in a data-driven manner. The proposed CNN architecture maintains differentiability and is end-to-end trainable. The proposed scheme is shown to clearly outperform other current techniques to achieve reliable correspondences across disparate images. We demonstrated the performance of the proposed system both qualitatively and quantitatively on challenging datasets.

\vspace{-0.15cm}
{\small
\bibliographystyle{IEEEtran}
\bibliography{bibliography_e2e.bib}
}
\end{document}